\documentclass[conference,a4paper]{IEEEtran}
\IEEEoverridecommandlockouts

\usepackage[hidelinks]{hyperref}
\usepackage[cmex10]{amsmath}
\usepackage{amssymb,amsfonts}
\usepackage{dblfloatfix}

\usepackage[ruled,vlined]{algorithm2e}
\usepackage{graphicx}
\graphicspath{{Figures/PDF/}{Figures/PNG/}}

\usepackage{booktabs}
\usepackage{siunitx}
\usepackage[numbers,compress]{natbib}
\usepackage{texnames}
\usepackage{bm,bbm}
\usepackage{orcidlink}

\begin{document}

\title{HQ-UNet: A Hybrid Quantum-Classical U-Net with a Quantum Bottleneck for Remote Sensing Image Segmentation}

\author{
    \IEEEauthorblockN{
        Md Aminur Hossain\orcidlink{0009-0003-6357-7480}$^{1,*}$, 
        Ayush V. Patel\orcidlink{0000-0002-8921-7091}$^{1,*}$, 
        Ikshwaku Vanani\orcidlink{0009-0006-1688-5562}$^{2}$,
        Biplab Banerjee\orcidlink{0000-0001-8371-8138}$^{3}$
    }
    \\
    \IEEEauthorblockA{
        $^{1}$Space Applications Centre, Indian Space Research Organisation (ISRO), India\\
        $^{2}$New-Brunswick Honors College, Rutgers University, New Jersey\\
        $^{3}$Centre of Studies in Resources Engineering, Indian Institute of Technology Bombay
    }
    \thanks{$^*$MA Hossain and AV Patel—Authors contributed equally to this research.}
}

\maketitle 

\maketitle
\begin{abstract}

Semantic segmentation in remote sensing is commonly addressed using classical deep learning architectures such as U-Net, which require a large number of parameters to model complex spatial relationships. Quantum machine learning (QML) provides an alternative representation paradigm by mapping classical features into quantum states, but its direct application to high-dimensional images remains challenging under near-term quantum hardware constraints. In this work, we propose HQ-UNet, a hybrid quantum–classical U-Net architecture that integrates a compact parameterized quantum circuit at the bottleneck of a classical U-Net. The proposed design uses a non-pooling quantum convolutional module to enrich highly compressed encoder features before decoding, while keeping the quantum component shallow and parameter-efficient. Experiments on the LandCover.ai dataset show that HQ-UNet achieves a mean IoU of 0.8050 and an overall accuracy of 94.76\%, outperforming the classical U-Net baseline. These results suggest that compact quantum bottlenecks can enhance feature representation for remote sensing image segmentation under near-term quantum constraints. This highlights the potential of hybrid quantum–classical designs as a promising direction for parameter-efficient dense prediction in Earth observation.
\end{abstract}

\section{Introduction}

Semantic segmentation assigns a semantic class label to each pixel in an image. It has become an essential part of many modern remote sensing analyses, allowing for a broad range of applications, including but not limited to urban planning and environmental monitoring~\cite{guo2018semantic, martins2021semantic}. While classical architectures based on deep learning, such as U-Net \cite{ronneberger2015u}, provide a strong benchmark for performance, these models tend to be very large (on the order of millions of parameters) due to their need to model the complex relationships between pixels in a spatially coherent fashion. Thus, they incur high computational costs and raise questions about the theoretical limits of classical models.

Quantum machine learning (QML) offers a new paradigm that leverages quantum principles like superposition and entanglement to process data through alternative representation spaces, enabling more efficient learning from high-dimensional data~\cite{biamonte2017quantum, schuld2015introduction, sebastianelli2026quantum}. The basic assumption of QML is that, by mapping classical data into a quantum state, a quantum circuit can provide an alternative feature transformation that helps capture complex correlations in compact latent representations. Unfortunately, due to the nature of Noisy Intermediate-Scale Quantum (NISQ)~\cite{wang2024comprehensive} hardware and its limited number of available qubits, applying QML to high-dimensional images will require a new hybrid quantum-classical approach~\cite{hossain2026qmcnet}. Specifically, a small quantum circuit will act as a specialized co-processor alongside classical computing resources.

In this work, we introduce HQ-UNet (Hybrid Quantum U-Net), a novel architecture that combines quantum and classical components for end-to-end segmentation tasks where at the lowest point of compression, a Quantum Convolutional Neural Network (QCNN) is employed to extract the deepest features. While foundational principles for quantum computing models, such as the QCNN \cite{cong2019quantum}, have been explored, the design of compact quantum bottlenecks for end-to-end remote sensing semantic segmentation remains relatively underexplored. HQ-UNet advances our understanding of how quantum architectures can enhance deep learning models for complex tasks. The key contributions of this work are as follows:

\begin{itemize}
    \item \textbf{Hybrid U-Net Architectural Design:} HQ-UNet integrates a non-pooling QCNN at the bottleneck of a classical U-Net, providing a higher-dimensional feature representation that enables effective reconstruction of spatial structures in the decoder.
    
    \item \textbf{Spectral-Aware Quantum Encoding:} A spectral-aware encoding scheme maps multi-channel encoder features to qubit states using parameterized $R$ rotations ($RX, RY, RZ$), enabling richer and more expressive quantum feature representations.
    
    \item \textbf{2D Quanvolution:} The proposed QCNN employs separable, parameterized filters along horizontal and vertical directions, adapting classical separable convolutions for efficient processing of 2D spatial data.
\end{itemize}

\section{Literature Review}

HQ-UNet builds on advances in two areas of research: the development of the U-Net deep learning model that is widely used today to provide semantic segmentation of images, and the growing field of QML. In this review, we set the stage for our research work by looking at some of the key classical and quantum-inspired architectures that have influenced our hybrid approach to building our new model.

U-Net, introduced by Ronneberger et al.~\cite{ronneberger2015u}, has become an important building block for modern segmentation networks. It is built in a symmetric fashion where the encoder and decoder are mirrored versions of each other, and the skip connections allow for merging the deep abstract knowledge from the encoder with the fine resolution images being decoded. This design has proven to be very helpful, producing state-of-the-art performance metrics on medical imaging datasets, and has been adapted for a variety of remote sensing applications. D-LinkNet~\cite{zhou2018d}, for instance, modifies the U-Net architecture by using a pretrained ResNet~\cite{he2016deep} encoder and has achieved very high rankings in recent competitions such as the DeepGlobe~\cite{demir2018deepglobe} satellite image road extraction challenge, which further establishes the U-Net architecture to be the standard classical architecture choice in this field.

Research into quantum machine learning (QML) has been able to develop a base for what will eventually be a quantum-enhanced model, as evidenced by the Quantum Convolutional Neural Network (QCNN)~\cite{cong2019quantum, aburaed2026review, devadas2025quantum}. The QCNN was defined as a hierarchical quantum circuit composed of alternating convolutional and pooling layers; in their work, the authors demonstrated that quantum circuits were capable of learning how to perform complex data classifications from a quantum circuit. Further progress on the QCNN was made by Henderson et al. \cite{henderson2020quanvolutional}, where an alternative framework for the QCNN was developed. The authors proposed an approach that combined a classical architecture with a small, parameterized quantum circuit that serves as a "quanvolutional filter." In their research, the authors demonstrated that small quantum circuits could extract relevant features from classical datasets.

The primary application of these principles lies in the use of a "quantum bottleneck" architecture. Quantum circuits are located within a deep classical network at the point of highest compression. This hybrid approach has been experimentally tested and confirmed to work in other complex application areas. For example, Li et al.~\cite{li2022image} explored a hybrid quantum-classical convolutional network for remote sensing image classification, showing the feasibility of combining quantum circuits with classical feature extractors. Recent work on data-aware quantum representations supports the use of compact quantum modules for Earth observation tasks~\cite{hossain2026qmcnet}. However, one of the primary challenges facing any deep quantum model is the "barren plateau" problem, which can lead to exponentially vanishing gradients and make training difficult~\cite{pesah2021absence, wang2024comprehensive}. 

For segmentation tasks, researchers have employed quantum circuits at the bottleneck present between the encoder and decoder blocks of U-Net to extract complex features at the highest point of compression \cite{jain2025qufex, wang2025qc, halab2025qu, de2024towards, hossain2026hqf}. Therefore, we have strong theoretical justifications for the use of the hybrid approach combined with a small and shallow quantum component. Our goal is to use a custom non-pooling QCNN within the bottleneck of an optimally efficient U-Net architecture and accomplish extending the benefits of quantum bottlenecks from simple classification tasks to more complex end-to-end semantic segmentation. However, compact non-pooling QCNN bottlenecks for end-to-end remote sensing semantic segmentation remain relatively underexplored.

\section{Proposed Methodology}

HQ-UNet integrates a custom quantum circuit into a U-Net-based encoder-decoder framework~\cite {ronneberger2015u} for end-to-end semantic segmentation. This section describes the data preparation pipeline and the proposed new hybrid architecture. A high-level overview of HQ-UNet is shown in Fig.~\ref{fig:architecture}.

\subsection{Dataset and Pre-processing}
The LandCover.ai dataset \cite{boguszewski2021landcover} consists of aerial RGB orthophotos of Poland with a spatial resolution of approximately 25–50 cm/pixel. It provides pixel-level annotations for five classes: Building, Woodland, Water, Road, and Background. The original GeoTIFF files are large, approximately 9000 × 9500 pixels. Therefore, a multi-stage pre-processing pipeline was developed. First, each image and its corresponding mask were cropped into 512 × 512 tiles and saved as lossless PNGs. Dynamic random sampling was then used to extract normalized 128 × 128 patches before augmentation.

\subsection{Hybrid Quantum U-Net (HQ-UNet) Architecture}
The established U-Net structure as an Encoder/Decoder Network is replicated here, with one exception: the deepest classical convolutional block (the bottleneck) of the U-Net has been replaced with a quantum processor. This has allowed the majority of upscaling and the creation of large-scale spatial features to be performed via the classical network, which is highly efficient, while letting the quantum circuit deal only with highly compressed, abstract features.

\begin{figure}[!ht]
    \centering
    \includegraphics[width=\columnwidth]{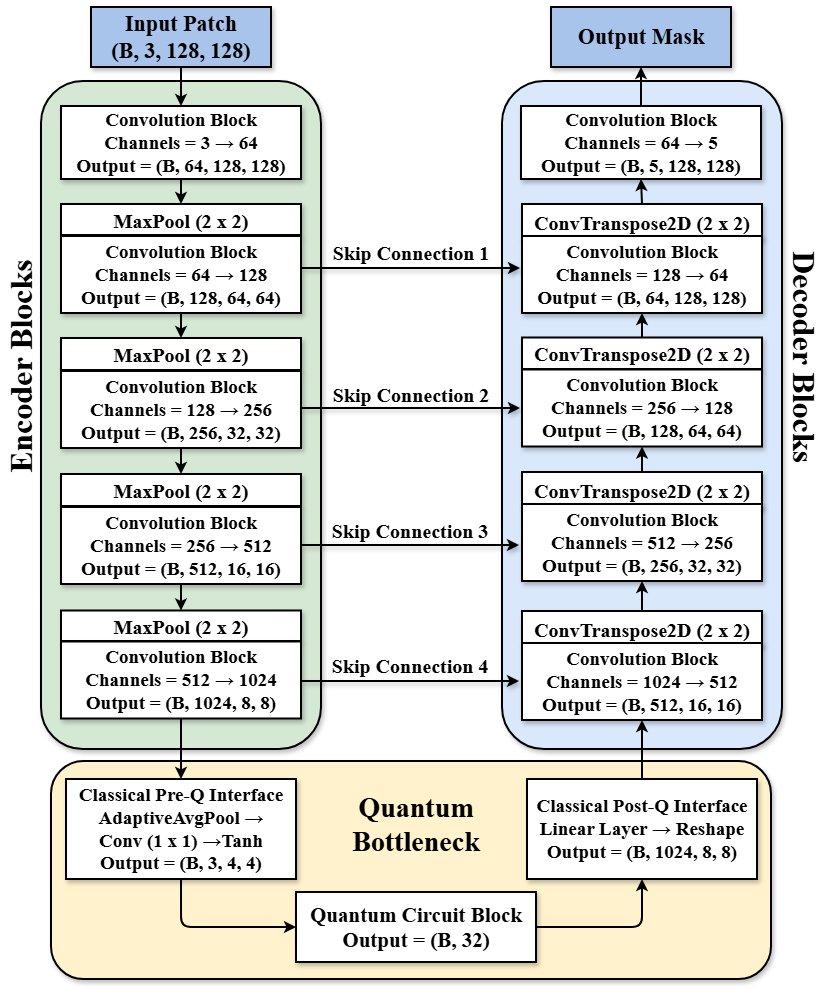}
    \caption{Overview of the proposed HQ-UNet architecture with a quantum bottleneck.}
    \label{fig:architecture}
\end{figure}

\subsubsection{Classical Encoder and Decoder} The classical component of this model is built using a depthwise separable convolution, which enables the convolution to be completed in a parameter-efficient manner. The standard form of a Convolution is now broken down into two smaller steps, depthwise and pointwise (using $1\times1$ convolutions). The encoder includes four separate down blocks, each consisting of a MaxPool2d layer followed by a DoubleConv block (2 depthwise separable convolutions coupled with BatchNorm and ReLU functions). Similar to the encoder path, the decoder path has four separate Up blocks. Each Up block will take the input from the previous Up block and use a ConvTranspose2d operation to get it back up to the input resolution, and then concatenate this up-sampled output with the corresponding high-resolution feature map from the encoder side through a skip connection, then refine these two features using the DoubleConv block, and finally, through a $1\times1$ convolution channel produce a prediction (number of classes).

\begin{figure}[!ht]
    \centering
    \includegraphics[width=\columnwidth]{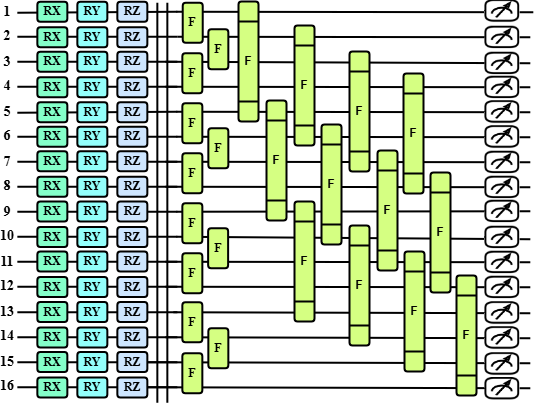}
    \caption{Quantum circuit implementing the proposed non-pooling QCNN used as the bottleneck in HQ-UNet. $F$ denotes the 2-qubit convolutional filter, where two RY gates are applied before a CNOT operation, and two additional RY gates are applied afterward.}
    \label{fig:quantum_circuit}
\end{figure}

\subsubsection{The Quantum Bottleneck} The quantum bottleneck, illustrated in Fig.~\ref{fig:architecture}, enriches the highly abstracted features produced by the encoder by operating on the final encoder feature map through a compact quantum-state representation. As shown in the quantum circuit of Fig.~\ref{fig:quantum_circuit}, this process is realized through three sequential stages, described below:
\begin{itemize}
    \item Classical Pre-Q Interface: The features generated by the final encoder layer are spatially reduced to a 4 × 4 grid for 16 qubits using AdaptiveAvgPool2d. Then, there is a $1\times1$ convolution which produces three separate channels. The channels are then modified to have values in the range of $[-1, 1]$ by passing through a Tanh activation function.
    \item Quantum Circuit Block: Once all three features are available for a given position (i.e., $f_1, f_2, f_3$), the features are then mapped to their respective qubit states based on the use of a spectral-aware encoding scheme of a sequence of $RX, RY, RZ$ rotation gates~\cite{du2025quantum}. The non-pooling QCNN performs a 2D separable quanvolution operation using a 2-qubit convolutional filter \(F\) defined in Eq.~\eqref{eq:quantum_circuit}, to model spatial correlations across the encoded quantum features.
    \begin{equation}
        \begin{split}
            F(\boldsymbol{\theta}, \boldsymbol{\phi}) = {} & [RY(\phi_1) \otimes RY(\phi_2)] \\
            & \cdot \text{CNOT} \cdot [RY(\theta_1) \otimes RY(\theta_2)]
            \end{split}
            \label{eq:quantum_circuit}
    \end{equation}
    This includes a two-pass process where the first pass occurs horizontally across multiple rows by applying the $F_R(\theta, \phi)$ parameterized 2-qubit filter across all rows; and the second pass occurs vertically down the multiple columns, applying the independent $F_C(\theta, \phi)$ filter down each column. This shared-parameter design enables the circuit to model non-local correlations among the encoded features while maintaining parameter efficiency.
    \item Classical Post-Q Interface: It extracts a robust representation by applying a multi-basis measurement to calculate the Pauli-Z and Pauli-X expectation values for every qubit. The resulting measurements form a $2N_q$-dimensional feature vector, where $N_q$ is the number of qubits. A linear layer then maps the quantum feature vector back to a classical feature map with the required channel depth and spatial dimensions before passing it to the first decoder block.
\end{itemize}

All quantum operations were implemented using a noiseless quantum simulator, while keeping the circuit depth and qubit count small to remain compatible with near-term NISQ constraints and practical hardware limitations.

\section{Results and Analysis}
This section presents the quantitative and qualitative evaluation of the proposed HQ-UNet on the LandCover.ai dataset. The performance of HQ-UNet is compared with representative hybrid quantum models, a lightweight classical CNN, and standard U-Net-based architectures to assess its effectiveness for remote sensing image segmentation.

\subsection{Quantitative Results}
The two primary metrics that we evaluated on were mean Intersection Over Union (mIoU) and overall pixel accuracy (OA) \cite{wang2023revisiting}. Mean Intersection over Union (mIoU) measures the class-wise overlap between predicted and reference masks, while overall accuracy (OA) measures the percentage of correctly classified pixels. All comparative performances can be found in Table~\ref{tab:results}.

\begin{table}[!t]
\centering
\caption{Comparative Performance on the landCover.ai Test Set}
\label{tab:results}
\begin{tabular}{llcc}
\toprule
\textbf{Source} & \textbf{Model} & \textbf{mIoU} & \textbf{OA(\%)} \\
\midrule
Fan et al. \cite{fan2024land} & FQCNN & 0.2000 & 53.26 \\
Fan et al. \cite{fan2024land} & MQCNN & 0.1520 & 39.03 \\
Kumar et al. \cite{kumar2024remote} & CNN & 0.1500 & 45.87 \\
Kumar et al. \cite{kumar2024remote} & COQCNN & 0.1280 & 36.65 \\
Ronneberger et al. \cite{ronneberger2015u} & UNet & 0.6451 & 82.43 \\
Zhou et al. \cite{zhou2018unet++} & UNet++ & 0.6553 & 70.89 \\
Abdani et al. \cite{abdani2020u} & UNet SPP & 0.6920 & 71.27 \\
Priyanka et al. \cite{priyanka2022diresunet} & DIResUNet & 0.7522 & 87.05 \\
\midrule
\textbf{Ours} & \textbf{HQ-UNet} & \textbf{0.8050} & \textbf{94.76} \\
\bottomrule
\end{tabular}
\end{table}

HQ-UNet achieves an mIoU of 0.8050 and OA of 94.76\%, outperforming the classical U-Net baseline by 0.1599 mIoU and 12.33\% OA. It also shows a clear margin over existing hybrid quantum baselines, such as FQCNN with an mIoU of 0.2000. These results demonstrate the effectiveness of the proposed quantum bottleneck design.

\subsection{Analysis and Discussion}
The strong performance of HQ-UNet arises from the effective integration of a quantum bottleneck into a robust U-Net framework with skip connections, thereby preserving spatial accuracy in the reconstructed segmentation masks. Building upon the classical U-Net baseline (mIoU 0.6451), HQ-UNet achieves improved performance by replacing the conventional bottleneck with a custom non-pooling QCNN.

The proposed quantum bottleneck enables enhanced modeling of higher-level feature correlations while introducing only a small number of trainable quantum parameters, contributing to improved parameter efficiency. In contrast, existing hybrid models often suffer from ineffective global encoding or insufficient architectural depth, limiting their performance on complex semantic segmentation tasks.

\subsection{Qualitative Analysis}
\begin{figure}[!ht]
    \centering
    \includegraphics[width=\columnwidth]{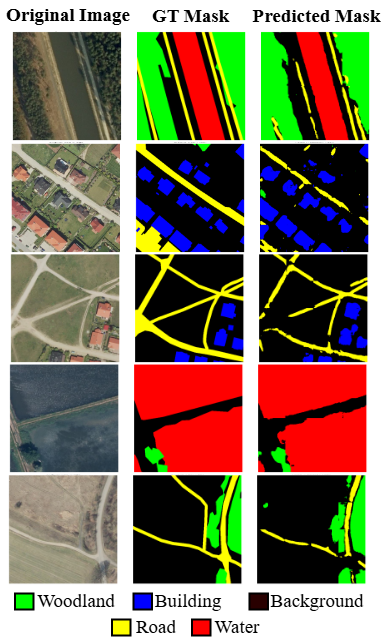}
    \caption{Qualitative segmentation results on the LandCover.ai dataset showing original images, ground-truth masks, and HQ-UNet predictions.}
    \label{fig:lcai_dataset}
\end{figure}

The qualitative results in Fig.~\ref{fig:lcai_dataset} support the quantitative findings. HQ-UNet produces coherent segmentation masks that closely follow the ground-truth labels. It captures fine linear structures such as roads, separates building footprints, and preserves clear boundaries between land-cover classes. Minor artifacts remain, including occasional confusion between Woodland and Water and slight thickening of road segments at complex intersections. However, these errors are limited and do not significantly affect the overall segmentation quality.

\section{Conclusion and Future Work}
This work introduced HQ-UNet, a hybrid quantum–classical architecture for remote sensing image semantic segmentation that integrates a non-pooling quantum convolutional neural network (QCNN) as a bottleneck within a classical U-Net framework. Experiments on the LandCover.ai dataset demonstrate that HQ-UNet achieves a mean Intersection over Union (mIoU) of 0.8050 and overall pixel accuracy of 94.76\%, outperforming the classical U-Net baseline. These results highlight the effectiveness of strategically embedding parameter-efficient quantum circuits to enrich deep feature representations in end-to-end segmentation tasks. Future studies will evaluate HQ-UNet on additional remote sensing and medical image datasets to assess its generalization capability. Ablation studies will also be conducted to quantify the contribution of each architectural component, including the quantum bottleneck and the encoding strategy.

\section*{Acknowledgments}
The authors would like to thank the Director of the Space Applications Centre (SAC), ISRO, for his encouragement and support. They are also grateful to the members of the Signal and Image Processing Area (SIPA) at SAC, Ahmedabad, India, for their valuable feedback on this research.
\bibliographystyle{IEEEbib}
\bibliography{references}

\end{document}